\documentclass[runningheads]{llncs}
\usepackage[T1]{fontenc}
\usepackage{graphicx}
\usepackage[caption=false]{subfig}
\usepackage{subcaption}
\usepackage{amsmath}
\usepackage{enumerate}
\usepackage{amssymb}
\usepackage{hyperref}
\usepackage{multirow}
\usepackage[ruled,vlined ]{algorithm2e}

\begin{document}
\title{A Novel Ranking Scheme for the Performance Analysis of Stochastic Optimization Algorithms using the Principles of Severity}
\titlerunning{Severity Based Ranking Scheme}
%
\author{Sowmya Chandrasekaran \orcidID{0000-0002-5304-6411} \and
Thomas Bartz-Beielstein \orcidID{0000-0002-5938-5158}
}
\authorrunning{F. Author et al.}
%
\institute{Institute for Data Science, Engineering, and Analytics, TH K\"oln, Steinmüllerallee 1, 51643 Gummersbach, Germany \\
\email{\{sowmya.chandrasekaran,thomas.bartz-beielstein\}@th-koeln.de).}
\\
}
\maketitle              
\begin{abstract}
Stochastic optimization algorithms have been successfully applied in several domains to find optimal solutions. Because of the ever-growing complexity of the integrated systems, novel stochastic algorithms are being proposed, which makes the task of the performance analysis of the algorithms extremely important.  In this paper, we provide a novel ranking scheme to rank the algorithms over multiple single-objective optimization problems. The results of the algorithms are compared using a robust bootstrapping-based hypothesis testing procedure that is based on the principles of severity. Analogous to the football league scoring scheme, we propose pairwise comparison of algorithms as in league competition. Each algorithm accumulates \emph{points} and a performance metric of how good or bad it performed against other algorithms analogous to goal differences metric in football league scoring system. The goal differences performance metric can not only be used as a tie-breaker but also be used to obtain a quantitative performance of each algorithm. The key novelty of the proposed ranking scheme is that it takes into account the performance of each algorithm considering the magnitude of the achieved performance improvement along with its practical relevance and does not have any distributional assumptions. To demonstrate the advantages of the proposed ranking scheme, we compare the expected run-time metrics of three hyperparameter optimization (HPO) procedures, namely, Irace, a mixed-integer parallel efficient global optimization (MIP-EGO), the mixed-integer evolution strategy (MIES), along with (1+1)EA and grid search(GS) on a genetic algorithm framework for Pseudo-Boolean Optimization (PBO) Suite of 25 problems. The proposed ranking scheme is compared to classical hypothesis testing and the analysis of the results shows that the results are comparable and our proposed ranking showcases many additional benefits. 

\end{abstract}
\section{Introduction}
Numerous new nature-inspired meta-heuristic algorithms are being proposed to solve various complex problems. This makes the analysis of the performances of the algorithms to a relevant set of problems an inevitable task. Generally, the performances of the stochastic optimization algorithms are evaluated based on solution quality or utilized budget \cite{bart20q}. Here, the solution quality measures how close the solution obtained by an algorithm is with respect to the global optimum or the best-known value. This is referred to as the \emph{fixed-budget} measure, where the achievable solution quality for a fixed budget is obtained. In the \emph{fixed-target} perspective, the time required by an algorithm to hit the desired solution quality is measured. The time required can be CPU time or function evaluations. Typically, the CPU time can be dependent on many factors like computing environment, hardware resources, etc. Hence, the Function Evaluations (FE) is considered as an alternative time measure, where the number of times the objective function evaluated is measured. To test the robustness of the algorithm's performances, the algorithms are tested under uncertainty, noise, etc. The scalability measures the ability of the algorithm as the dimension of the problem increases. Be it the fixed-target or fixed-budget measure, due to the stochastic nature of the algorithm, there exists randomness in the performance of algorithms. Executing the same algorithm repeatedly can produce different solutions for the same inputs. Hence, there is a need for rigorous analysis of the performances of stochastic optimization algorithms.

In recent years, descriptive analysis (e.g. mean, median, best, worst and standard deviation) has turned out to be necessary but not sufficient metrics in analysing the performances of the algorithms. Statistical analysis plays a crucial role in comparing the performances of the algorithms \cite{bart20q,carr11a,shil08a,chri04a,derr11a,garc09a}. 
A commonly used statistical tool over several years is hypothesis testing \cite{moor89a,neym28a}. In order to compare the performances of algorithms, the null hypothesis can be formulated as \emph{There is no statistically significant performance difference between a pair of algorithms} $vs$ the alternative hypothesis as \emph{ There exists a statistically significant performance difference between a pair of algorithms.} Hypothesis testing can be broadly classified into parametric and non-parametric tests. While the former assumes a specific type of probability distribution of the data and makes inferences about the parameters of the distribution, the latter do not make any explicit assumptions about the data or its underlying distributions. The non-parametric tests are used when the assumptions for the safe use of parametric tests are not met. In both procedures, the $p$-value  be the measure for deciding whether to retain or reject the null hypothesis. In \cite{chri04a,shil08a,derr11a,garc09a,garc10a,moli18a}, use cases to apply parametric or non-parametric tests to evaluate the performances of the meta-heuristic optimization algorithms are discussed. However, there exist pitfalls in using the hypothesis testing procedure as it can easily be misused and misinterpreted. Since the statistical significance is decided in the form of a yes or no fashion that is based only on the $p$-value, the hypothesis testing procedure is criticized as black and white thinking  \cite{bena17a}. Considering these criticisms, in \cite{asa16a}, the American Statistical Association (ASA) explains the scope of $p$-value, wherein it emphasizes on considering additional appropriate measures along with $p$-value for a scientific decision. In \cite{chan23a}, this issue is addressed using a measure, severity, which is a form of attained power \cite{mayo06a}. More precisely, a bootstrapping-based distribution-free robust statistical framework for the analysis of the performances of stochastic algorithms is proposed in \cite{chan23a}, where  both the statistical significance and also the practical relevance of the algorithms performances are measured. Also, the concept of severity is utilized for the analysis of the performances of hyper-parameter tuning in machine and deep learning algorithms \cite{bart23a}.

The purpose of this paper is to propose a novel ranking scheme, that ranks the algorithms in a robust statistical fashion based on their performances considering the statistical significance, practical significance and magnitude of the achieved performance improvement. The resulting ranking scheme is analogous to the football league ranking system which has both the \emph{points} scored and goal differences metric. Considering Multiple Algorithm Multiple Problem design (MAMP), each pairwise comparison of algorithms is treated similarly to a football game between two teams in a football league competition. The outcome of each game can be a \emph{win}, a \emph{draw}, or a \emph{loss} for each algorithm on each problem based on which the \emph{points} and \emph{goal difference} are obtained. To the best of our knowledge, this may be the first statistical ranking scheme, which takes into account the win or loss of an algorithm along with the magnitude of the corresponding win or loss, i.e, in terms of the positive or negative goal differences. At the end of the league competitions, the algorithms are ranked based on the \emph{points} and \emph{goal difference}(GD). The key contributions of the proposed ranking scheme are summarized below:
\begin{enumerate}
    \item The \emph{points} for each algorithm are evaluated considering the decision of the hypothesis testing, the statistical significance of the win or loss, and the practical relevance of the achieved win or loss. 
    \item The \emph{goal difference}, the magnitude of the corresponding win or loss, is used as a tie-breaker when two algorithms attain the same \emph{points}. The goal difference can be positive or negative based on the league competition outcome. 
    \item This ranking scheme sheds light on understanding the success or failure of an algorithm's performance in a robust measurable fashion while providing a quantitative understanding of the performances of each algorithm.
\end{enumerate}

The paper is organized as follows: Section \ref{sec:rw} explains the existing ranking frameworks to evaluate the performances of the optimization algorithms. Section \ref{sec:sev} explains the concept of severity. Section \ref{sec:proposed-tool} summarizes the proposed football based ranking scheme. In Section \ref{sec:casestudy}, a specific set of hyperparameter optimization techniques are evaluated for a family of genetic algorithms and are tested using the PBO Suite of 25 problems. Section \ref{sec:summary} concludes with a summary and outlook. \\

 \section{Related Works}\label{sec:rw}
 In order to rigorously benchmark, compare and analyze wide range of optimization algorithms, bench-marking tools like Iohprofiler are proposed \cite{doer20a}. The IOHexperimenter  component of Iohprofiler \cite{IOHe21a}, well-known bench-marking suites like Pseudo-Boolean Optimization (PBO) \cite{pbo20a}, Black-Box Bench marking (BBOB) can be generated to obtain the experimental data \cite{hans21a,hans09a}. The Iohanalyzer component provides statistical and graphical tools for analysis and visualization of the experimental data \cite{ioh22a}. In addition, several other bench-marking suites from the Special Session on Real Parameter Optimization organized at IEEE Congress on Evolutionary Computation (CEC) in the years 2013 \cite{cec13a}, 2014 \cite{cec13b}, 2015 \cite{cec14a}, and 2017 \cite{cec17a} are available. 
 
 There exist some ranking schemes in the literature which were proposed to compare the performances of the optimization algorithms as in \cite{efti17a,efti19b,efti19a,calv19a,mers10a,rojas22a}. In \cite{Niki14a}, an empirical chess rating system for evolutionary algorithms using Glicko-2 rating is proposed. Here, the evolutionary algorithms are treated as chess players, and a pairwise comparison of two algorithms is considered as one game. Each game outcome can be a \emph{win}, a \emph{draw}, or a \emph{loss}. At the end of the tournament, each algorithm is represented by rating(R), rating deviation (RD), and rating volatility($\sigma$). Different variants of this chess rating system are compared in \cite{vevc14a} and the Glicko-2 rating is identified to be more reliable. Despite not being statistically analyzed, this rating system suffers from other issues. Firstly, the ordering of the games affects the final rating, though it is randomly selected. Furthermore, the overlapping of the confidence intervals might lead to statistical inconsistency as explained in \cite{efti19b}. Finally, the magnitude of the win or loss is not considered in the rating system. 
 
 More recently, different variants of deep statistic-based comparison (DSC) tool have been proposed \cite{efti17a,efti19b,efti19a}. The advantage of this \cite{efti19a} ranking scheme is that it is based on the whole distribution rather than comparing mean or medians. In MAMP design, multiple pairwise comparisons of algorithms are performed using the non-parametric Kolmogorov-Smirnov or Anderson-Darling test and only the p-values determine the win or loss. Though the practical significance is addressed in DSC, the magnitude of the win or loss is not considered. Also, since the practical significance is directly included in the hypothesis formulation, the approach can be  more conservative. 
 
 In \cite{calv19a,rojas22a}, the statistical comparison of the performances of the evolutionary algorithms is performed using Bayesian inferences. Though the identification of prior probabilities is an issue \cite{gelm08a}, the ranking of the algorithms is only based on the Bayesian probability of an algorithm being the best performer. The magnitude of the performance differences cannot be obtained.
 
 In \cite{mers10a}, a consensus ranking system is proposed, where the performance of algorithms across multiple problems are aggregated to determine an overall ranking based on their collective performance. The scope and limitations of this ranking are well explained in \cite{Olaf23a}. 
 \section{Proposed Ranking Scheme}\label{sec:prs}
 \subsection{Concept of Severity}\label{sec:sev}
 In order to explain the concept of hypothesis testing and severity, let us assume Normal, Independent, and Identically Distributed (NIID) data\footnote{In the proposed ranking scheme, we do not assume the normality of the data. Here it is assumed to simplify the explanation of the concept and without the loss of generality, the concept can be adapted to the cases where the distribution is not known.}. 
 Let us consider algorithm $A$, say $\mathbf{a}=(a_1,\,\dots,\, a_n)$, representing the function evaluations required by Algorithm $A$ to achieve a specific target solution for $n$ runs. Similarly, algorithm $B$, say $\mathbf{b}=(b_1,\,\dots,\, b_n),$ represents the function evaluations required by Algorithm $B$ to achieve the same  target solution for $n$ runs. We evaluate the performance of the algorithms repeatedly for $n$ runs to handle the randomness in the evaluation metric and to obtain a reliable estimate of the metric.
 
 The hypothesis testing is performed as an upper tail test of the mean differences as it is an optimization minimization problem. The difference vector $\mathbf{x}$ can be defined as $\mathbf{x}=(x_1,\,\dots,\, x_n)$, where $x_i=a_i-b_i,\,\forall i = \{1,\,\dots,\,n\}$ and $\mathbf{\bar{x}}$ denote the mean of the vector $\mathbf{x}$. \\
$H_0:$ $B$ does not achieve less FE than $A$ $\implies \mathbf{\bar{x}} \leq 0$ $\implies$ Loss for $B$.\\
$H_1:$  $B$ achieves less FE than $A$ $ \implies  \mathbf{\bar{x}}>0$ $\implies$ Win for $B$.\\        
 \begin{align}\label{eq:np}
  H_0: &\mathbf{\bar{x}} \leq 0;\mathrm{ vs. } \;H_1: \mathbf{\bar{x}}>0,\\
   decision&=
    \begin{cases}
     \text{not-Reject}\, H_0 , & \text{if}\ \ d(\mathbf{X}) \leq u_{1-\alpha}, \nonumber\\
      \text{Reject}\, H_0, & \text{otherwise},
    \end{cases}
 \end{align}
 where $u_{1-\alpha}$ is the upper tail cut-off point of the normal distribution, which cuts the upper-tail probability of $\alpha$, and the test statistic $d(\mathbf{X})$ can be represented as   
 \begin{equation}
d(\mathbf{X}) = \frac{\bar X-\mu_0}{\sigma_x },
 \end{equation}  
 where standard error $\sigma_x=\frac{\sigma}{\sqrt{N}}$ and $\mu_0$\ is the hypothesized mean under $H_0$. If a test statistic is observed beyond the cut-off point, we reject the $H_0$ at a significance level $\alpha$. Here, the values for $\alpha$, $\beta$, and hence power ($1-\beta$) are pre specified before the experiment is performed.
 
\begin{figure}[!]
\centering
    \subfloat[Won Scenario 1 :severity to Reject H0] {{\includegraphics[width=2.25in]{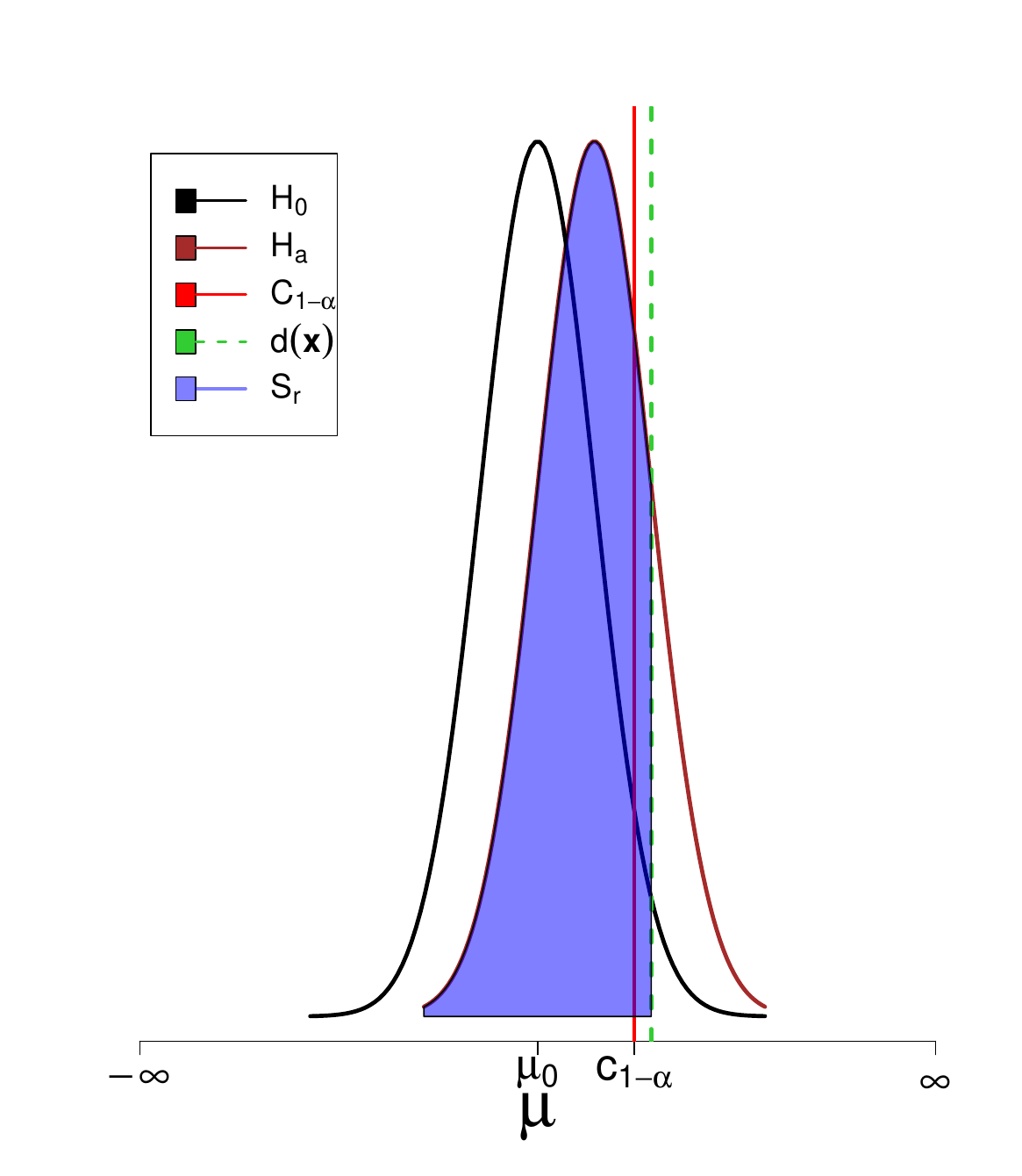} }}
    \hfill
    \subfloat[Won Scenario 2: severity to Reject H0] {{\includegraphics[width=2.25in]{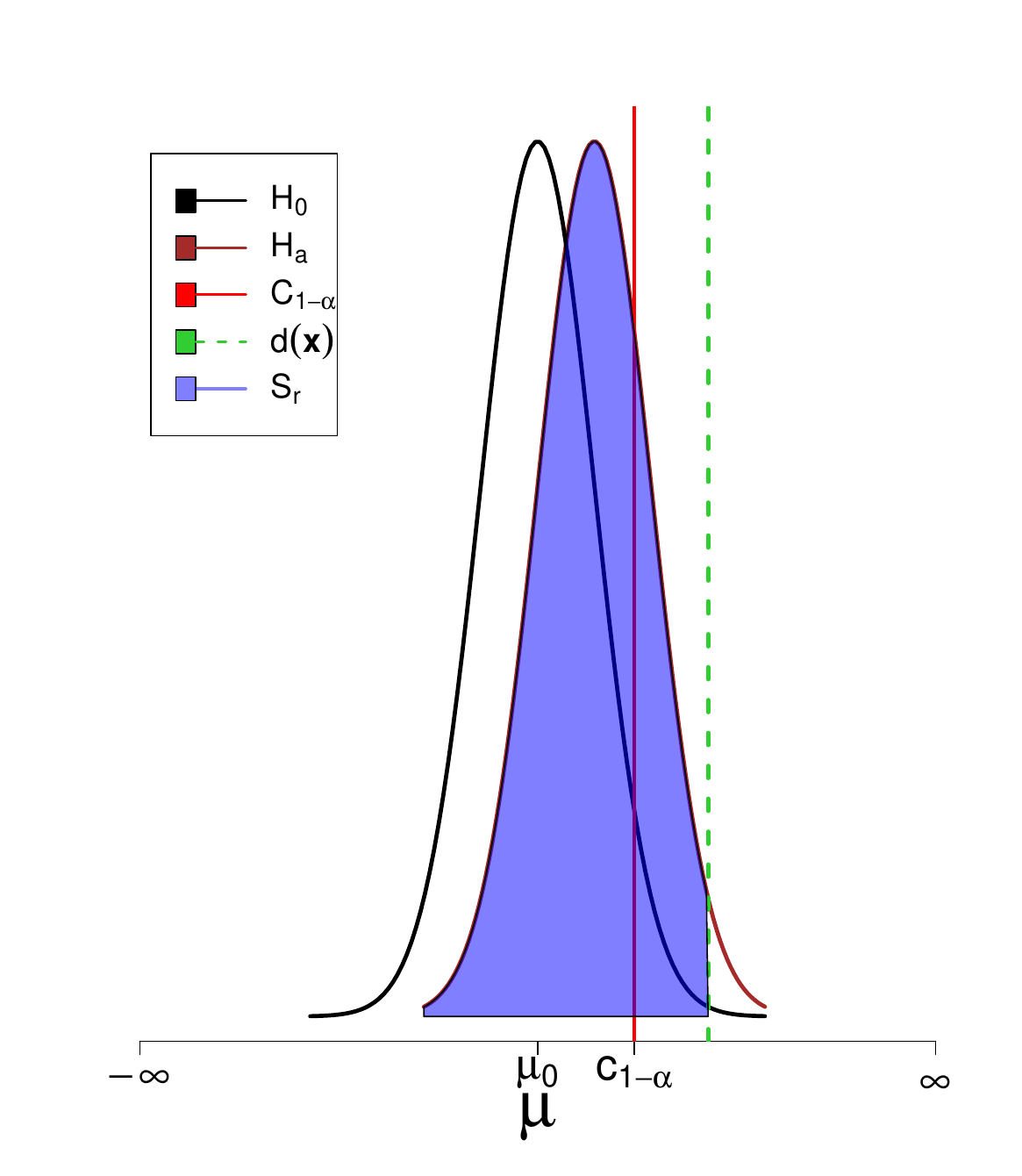} }}
    \caption{Illustration of two scenarios of $S_{r}$  under the alternate hypothesis. In both cases, the actual test statistic $d(\mathbf{x})$, falls outside the $u_{1-\alpha}$, the decision is to reject the null. The $S_{r}$ is the area under the $H_1$ that is within the $d(\mathbf{x})$ (area shaded in blue). Though in both cases, the decision is the same, severity sheds light in understanding the actual attained power of the test. In (a), less support for the decision won (area shaded in blue) as $d(\mathbf{x})$ is closer to the cut-off point and in (b), more support for the won (area shaded in blue) as $d(\mathbf{x})$ is way more from the cut-off point.}
\label{fig:sr}
  \end{figure}

 In the context of ranking the algorithms based on the results of parametric or non-parametric hypothesis testing, based only on the $p$-value, i.e, if a test statistic is observed beyond the cut-off point, the $H_0$ is rejected and hence $B$ wins and is declared of achieving less FE than $A$. Also, $B$ gains a point just based on this decision. This is criticized as black and white thinking.\\
 Severity, a form of attained power \cite{mayo06a}, is a probability analogous to the $p$‐value  under the alternative hypothesis rather than one under the null \cite{senn21a}. In case of win or loss, the magnitude of the performance improvement is measured in terms of severity as $S_{r}$ and $S_{nr}$ respectively. The loss $S_{nr}$ values increase monotonically from 0 to 1. The  won $S_{r}$ values decrease monotonically from 1 to 0. The closer the value is to 1, the more reliable the decision made with the hypothesis test. Generally, a severity value of 0.8 is considered reliable support. The differences between $\alpha$, $p$-value, power, and severity is provided in Table 1 in \cite{chan23a}.
The importance of severity representation of won is shown in Figure \ref{fig:sr}. In a similar fashion, the severity representation of loss can also be visualized. In Figure \ref{fig:sr}, only one value for the alternate hypothesis is visualized. However, in practice, we evaluate the severity for specific possible values under the $H_1$. The different values under the alternative hypothesis for which the compatibility is assessed will henceforth be called discrepancy, $\delta$. It measures how discrepant is the performance improvement when compared to the null improvement. In the context of algorithm ranking, since we consider the FE as the metric, we can evaluate if B won over A with 1000 FE or 10 FE, thereby quantization the magnitude of the victory. 
 \vspace{-1.5cm}
   \vspace{-1cm}
 \subsection{Proposed Algorithm}\label{sec:proposed-tool}
The football league scoring system has been designed to provide a fair and objective method for ranking teams based on their performances. Each football match between two teams, has three possible outcomes: win, loss or draw. Based on the outcome the teams get \emph{points}, 3 for winning, 1 for drawing and 0 for losing. The \emph{points} earned by each team from matches played over a season are cumulatively added and the final rankings for the season is obtained. In addition, the goal difference, which is the difference between the number of goals scored minus the number of goals conceded in matches is calculated. Goal difference serves as a tiebreaker if two or more teams have the same \emph{points}. 
\begin{algorithm}[H]
\ForEach { Pair-wise comparison $(A_i, A_j)_l \in \mathcal C$}
{
\KwData{ $(\mathbf{y}_i,\mathbf{y}_j)_l=((y_i^1,\,\dots,\, y_i^n),(y_j^1,\,\dots,\, y_j^n))_l$, $\alpha$, $S$,$n_b$, $\delta_{p}$}
\KwResult{ $\emph{points}$, $GD$}
Formulate Hypothesis $H_0:  \mathbf{\bar{x}}  \leq 0\;\mathrm{ vs. } \;H_1:  \mathbf{\bar{x}}  > 0$\\
Evaluate observed sample mean difference $t_{obs}=\overline{\mathbf{y}_i }-\overline{\mathbf{y}_j }$\\
 Combine $\mathbf{I} = \widehat{\mathbf{y}_i \mathbf{y}_j}$ \\
 \Repeat{$n_b$ times}{
 Draw a bootstrap sample of $2n$ observations with replacement from $\mathbf{I} $\\
 Let the mean of the first $n$ observation be $\overline{\mathbf{y}_i }^*$ \\
 Let the last $n$ observations be $\overline{\mathbf{y}_j }^*$  \\
  Evaluate $t^{*\text{bs}}=\overline{\mathbf{y}_i }^*-\overline{\mathbf{y}_j }^*$\\
  Evaluate $t_s^{*\text{bs}}=\overline{\mathbf{y}_i }^*-\overline{\mathbf{y}_j }^*-\delta$\\
 }
Calculate  $p\approx \frac{\#(t^{*\text{bs}} \geq t_{obs})}{n_b}$\\
Obtain $adjusted-p$ based on BH correction\\
 \eIf{ $ adjusted-p \leq \alpha$}{
\text{decision: Reject}  $H_0$ \\
 $\delta^*=\min_\delta  S - \frac{\#(t_s^{*\text{bs}} \leq t_{obs})}{n_b}$\\
 \eIf{  $\delta \leq \delta_{p}$}{
 $\emph{points}=1$\\
 $GD= 0$\\
 }
 {
 $\emph{points}=3$\\
 $GD= \lfloor\frac{\delta^*}{\delta_{p}}\rfloor$\\
 }
 }
{\text{decision: not-Reject} $ H_0$\\
 $\delta^*=\max_\delta  S - \frac{\#(t_s^{*\text{bs}} > t_{obs})}{n_b}$\\
 $\emph{points}=0$\\
 $GD= \lfloor\frac{\delta^*}{\delta_{p}}\rfloor$\\
}
}
Obtain cumulative $\emph{points}$ and $GD$\\
Rank based on $\emph{points}$\\
 \caption{ Proposed Ranking Scheme}\label{alg:alg_ranking}
\end{algorithm}
Analogous to the football league scoring scheme, we propose a novel ranking scheme for ordering the performances of several stochastic algorithms on a set of well-known bench-marking test functions ($\mathcal F$) as Algorithm \ref{alg:alg_ranking}. Let $\mathcal{A}:=\{A_i, \forall i \in \{1,\dots,k\}\}$ denote the set of all algorithms that must be ranked and $\mathcal{F}:=\{F_l, \forall l \in \{1,\dots,m\}\}$ denote the set of all test functions. An experiment is performed where each algorithm  $A_i\in \mathcal{A}$ is evaluated on each of the test functions  $F_l\in \mathcal F$ for $n$ runs. Let $\mathcal C$ represent set of all possible pair-wise algorithm comparisons,  $\mathcal Y$ represent the set of all corresponding solutions and are defined as
 \begin{align}\label{eq:ac}
 \mathcal{C}&:=\big\{(A_i,A_j)_l, \forall i, j \in \{1,\dots,k\}, i \neq j,\forall l\in \{1,\dots,m\} \big\},\\
 \mathcal{Y}&:=\big\{(\mathbf y_i,\mathbf y_j)_l, \forall i, j \in \{1,\dots,k\}, i \neq j,\forall l\in \{1,\dots,m\} \big\},
  \end{align}
where $\mathbf{y}_i \in\mathbb{R}^n$ denotes the solution of $i^\mathrm{th}$ algorithm $A_i$ for $n$ runs for a given function in $\mathcal{F}$. 
The ranking scheme requires each pairwise comparison defined in $\mathcal C$ and set $\mathcal{Y}$ should be obtained and this results in a total of  $k\times k-1\times m$ comparisons. For each $(A_i,A_j)_l\in\mathcal{C}$, bootstrapping-based t-test is performed. Sampling with replacement is done to attain a better estimate of the metric. This procedure also eliminates the dependence of the outcome on the ordering of the optimization runs. For a given value of $i, j \in \{1,\dots,k\}, \,l\in \{1,\dots,m\} ,i\neq j$, the following steps are performed.
\begin{enumerate}\label{steps_involved}
    \item Merge the results of the algorithms to obtain $\mathbf{I}:=\widehat{\mathbf{y}_i\mathbf{y}_j} $ of size $2n$ and compute the observed sample mean difference $t_{obs}:=\mathbf{\overline{y}}_i-\mathbf{\overline{y}}_j$.
    \item Draw a bootstrap sample of $2n$ observations with replacement from $\mathbf{I} $ and evaluate the bootstrap test statistic $t^{*\text{bs}}=\mathbf{\overline{y}}^*_i-\mathbf{\overline{y}}^*_j$. 
    \item Repeat the procedure based on the bootstrapping re-sample size $n_b$, and estimate $p$-value as the number of times the bootstrap test statistic $t^{*\text{bs}}$ was found to be greater than the $t_{obs}$ in $n_b$ samples. I.e., $p\approx \frac{\#(t^{*\text{bs}} \geq t_{obs})}{n_b}$
    \item Adjust the $p$-value using Benjamini-Hochberg (BH) correction \cite{Benj95a}.
   \item  When the adjusted $p$-value is found to be significant, i.e., less than the specified $\alpha$, then reject $H_0$, else do not reject $H_0$.  Based on the decision and the chosen  severity requirement, $S$, obtain the supported $\delta$. 
\end{enumerate}
The significance level $\alpha$, the desired severity, $S\in [0,\,1]$, (same as desired power) and the practically relevant $\delta_{p}$ should be chosen by the user. As the power of the test is usually chosen as 80 percent or 95 percent based on the problem domain, similarly, the recommended severity is  chosen at 80 percent or 95 percent and $\alpha$ of 95 percent. Based on the performance metric and the problem domain, to identify if the achieved performance improvement is better than the practical relevance, the $\delta_{p}$ shall be chosen carefully. In case of FE as the metric where the available budget is 100000 FEs, minimum of 100 FE improvement at desired severity can be considered as a practically relevant improvement. In case of CPU time as the metric, depending on the application, meaningful time can be chosen. E.g., for an optimization algorithm implemented in an autonomous car, this value can be in centiseconds, and for offline scheduling problems, it can be in several minutes to hours. The supported $\delta$ obtained from the algorithm gives a measure of change in statistics required to achieve the expected severity. The rounded-down value of the ratio of $\delta$ and $\delta_p$ provides the goal difference metric in each match, thereby quantifying the size of a win or a loss.

The \emph{points} scored for each algorithm are similar to the football league scoring system and are obtained based on the following criteria:
\begin{itemize}
    \item the decision of the bootstrapped hypothesis testing.
    \item the statistical significance of the win or loss, i.e. supported $\delta$ at desired severity
    \item the practical relevance of the win or loss, whether, supported $\delta< \delta_{p}$ or $\delta > \delta_{p}$ 
\end{itemize}
The \emph{goal difference} measures how much the performance is better/worse in terms of practical relevance. The \emph{points} and \emph{goal difference} are calculated as 

\begin{align}\label{eq:score}
  Outcome=
    \begin{cases}
     \text{\emph{points}=3, GD=$\lfloor\delta / \delta_{p}\rfloor > 0$}, \,  \text{if} \ \ \text{Reject}\, H_0 \, \ \text{and}& \text{if}\ \delta > \delta_{p},
     \nonumber\\
      \text{\emph{points}=1, GD=$\lfloor\delta / \delta_{p}\rfloor = 0$} , \, \text{if} \ \ \text{Reject}\, H_0 \ \text{and}  & \text{if}\ \ \delta < \delta_{p},
  \nonumber\\
      \text{\emph{points}=0, GD=$\lfloor\delta / \delta_{p} \rfloor\leq 0$} , \ \ \text{if} \ \ \text{not-Reject}\, H_0.
    \end{cases}
 \end{align}
Upon completion of all pairwise comparisons in $\mathcal C$, the cumulative \emph{points} and the cumulative goal differences are calculated and the algorithms are ranked based on the \emph{points}. In addition, the mean and standard deviations of the \emph{points} for each algorithms among all functions can be obtained. The points awarded to algorithms that performed well with practical significance are weighted more than twice (three times to be precise) the points awarded to algorithms that performed only statistically significant. This takes into account the fact that practical significance is more relevant in real-world applications and therefore provides a better estimate of the overall performance of the algorithms. On the other hand, weighting practical significance more than three times may skew the results more towards practical significance and therefore statistical significance will have less influence on the final result. The resulting scheme demonstrates the desired symmetry property, i.e, if Algorithm $A_i$ outperforms $A_j$, then $A_j$ did not outperform $A_i$ is also true.
\section{Case study}\label{sec:casestudy}
We compare the expected run time metrics of three hyper parameter optimization techniques along with (1+1)EA and grid search for a family of genetic algorithms on PBO Suite of 25 problems obtained from \cite{ye22a}. The 25 PBO functions include from onemax, leadingones, a linear function with harmonic weights, various W-model-transformes of onemax and leadingones, low autocorrelation binary sequences, ising models, maximum independent vertex set, N-queens problems. The compared HPO techniques include Irace \cite{irace16a}, a mixed-integer parallel efficient global optimization \cite{mipego19a}, the mixed-integer evolution strategy \cite{mies13a}. 

As (1+1)EA has shown good performance for PBO in \cite{doer20a} it is considered a baseline. The goal is analysing the impact of mutation, crossover, and its combination on a family of $(\mu+\lambda)$ GA algorithms, which results in four tuning parameters: Parent population size, $\mu \in [100]$, Offspring population size ,$\lambda \in [100]$,  mutation rate, $P_m \in [.005, .5]$, cross over probability  $P_c \in [0, 1]$. Each of the HPO techniques is allocated a budget of 5000 target runs, where each target run refers to 10 independent runs of the $(\mu+\lambda)$ GA configuration suggestion by the HPO techniques. Two different performance metrics are considered, namely, minimizing the expected runtime (ERT) and maximizing the Area under the empirical CDF curve of running times (AUC). This results in 9 algorithms to be compared: (1+1)EA, GS.AUC, GS.ERT, Irace.AUC, Irace.ERT, MIES.AUC, MIES.ERT, MIP.EGO.AUC, MIP.EGO.ERT. The $(\mu+\lambda)$ GA configurations provided by each of the HPO technique are evaluated for a budget of 50000 function evaluations and the ERT values are obtained for the PBO problems with respect to the targets defined in Table 1 in \cite{ye22a}. And for AUC, set of 100 equally spaced targets ranging from 0 to the targets defined in Table 1 in \cite{ye22a}  is considered. Upon identification of the best tuning parameters by each HPO technique, 100 independent runs for these tuned settings are performed and used for further analysis. For each of the algorithm, results of these 100 runs at desired target defined in Table 1 in \cite{ye22a} is used as performance data of this case study and is obtained using IOHanalyzer \cite{ioh22a} as explained in \cite{ye22a}. In all runs, values are capped at the budget 50000 function evaluations if the algorithm cannot ﬁnd the target.
The experimental setup for our ranking scheme is as follows: the significance level $\alpha$ is chosen as 0.05, and the desired severity is at recommended value of 0.8. Since the function evaluation is the performance metric in our fixed-target perspective and the total allocated budget is 50000 function evaluations, the practically significant performance improvement, $\delta_p$, can be chosen as minimum of 500 function evaluations. The re-sample size $n_b$ is chosen as 10000. The classical bootstrapped-based hypothesis testing (HT) is also performed with $\alpha$ as 0.05 and $n_b$ is chosen as 10000. Each algorithm scores one point for the decision Reject$H_0$ or zero otherwise. The resulting ranking results are shown in Table \ref{tab:result_summary}.

Without going into the details of the performance of the algorithms for each function, we can understand the quantitative performance of each algorithm from the table provided. For example, (1+1)EA clearly outperforms all the other algorithms and tops the table. However, looking at the GD metric, it can be observed that (1+1)EA has performed really poorly for the cases where it failed to win leading to a big negative goal difference at the end. Though MIES.ERT, MIES.AUC, Irace.AUC lags behind (1+1)EA in the \emph{points}, they have not performed really poorly for the cases it lost against other algorithms. If one would like to minimize the worst-case scenarios to achieve robustness, MIES.ERT can be chosen to be applied instead of (1+1)EA. In this case, we may not achieve the optimum the fastest. However, the results will not be the slowest for some classes of problems either.

\begin{figure}[h]
  \centering
   \subfloat[]{ \includegraphics[width=6cm]{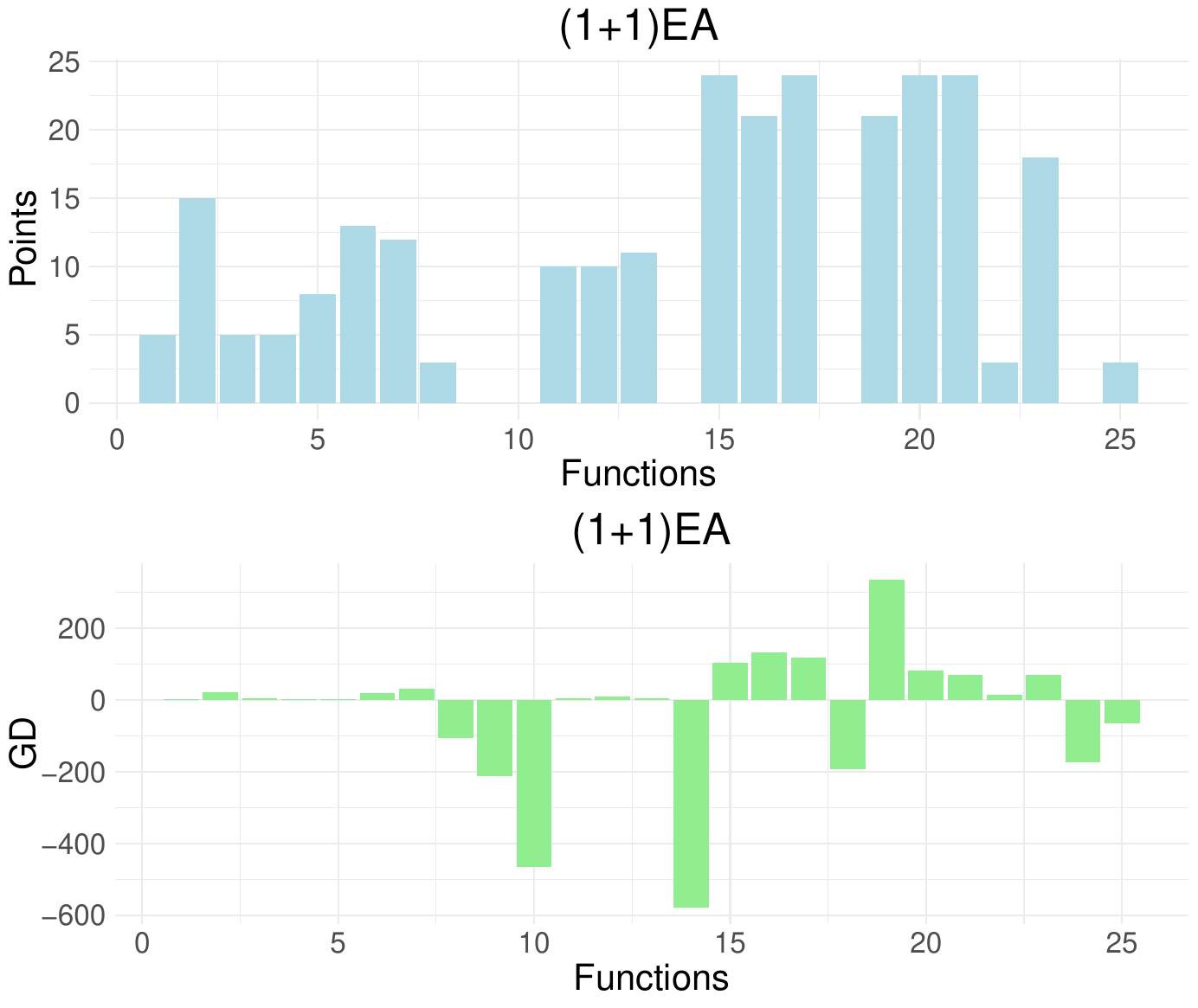}}
     \subfloat[]{ \includegraphics[width=6cm]{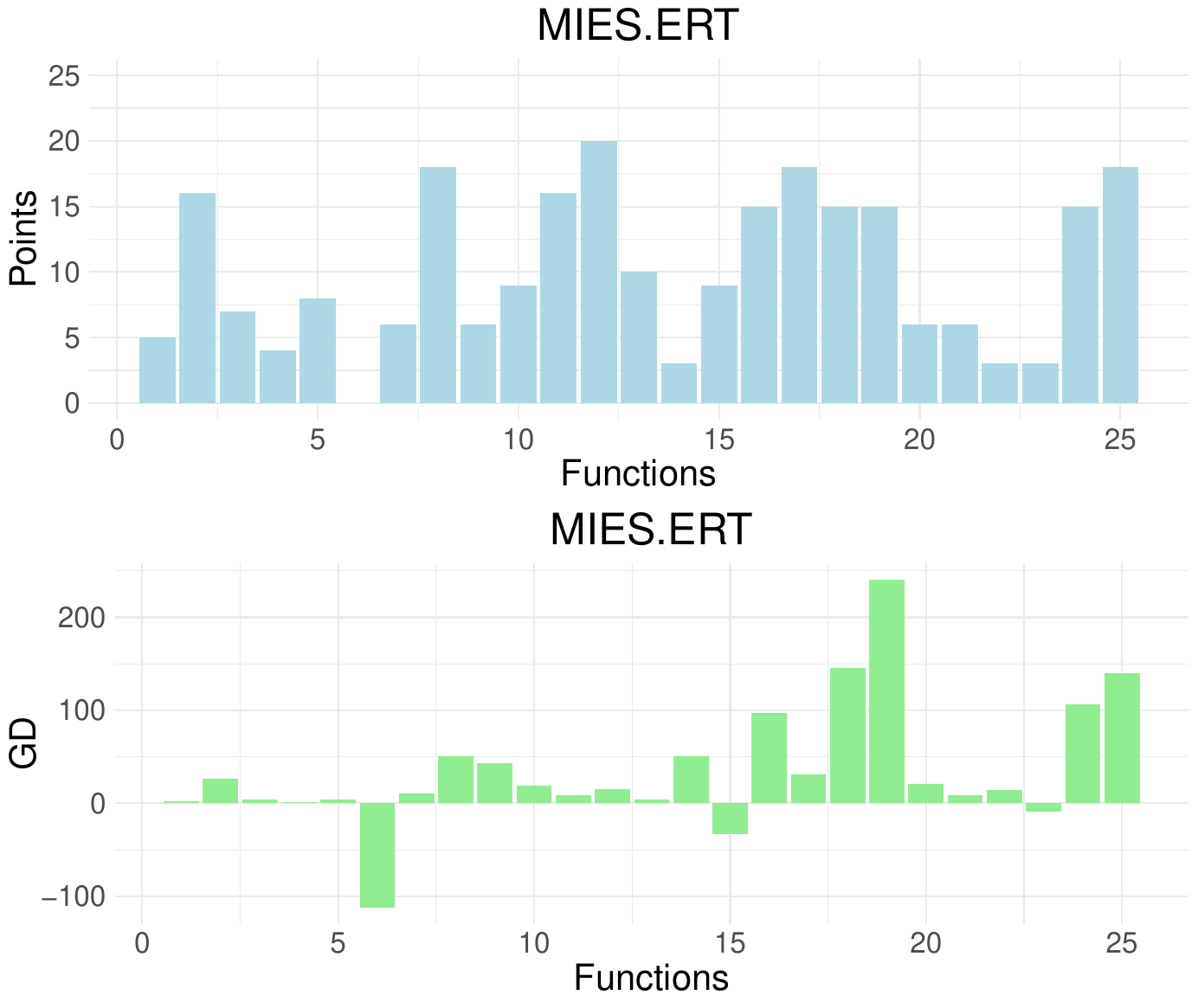}}
      \caption{Function-wise ranking metrics of the Rank 1:(1+1)EA and Rank 2: MIES-ERT Algorithms}
      \label{fig:eavsmies}
\end{figure}

The (1+1)EA algorithm was outperforming MIES.ERT for functions 6,7, 15-17, 19-21 and 23 as shown in Figure \ref{fig:eavsmies}. However, the positive GD in these functions was very low. For functions 8-10,14,18, and 24, where the (1+1)EA could not outperform, the GD was negative. The MIES.ERT was able to outperform (1+1)EA in functions 2,3,8-14,18,24, and 25 with positive GD for the majority of the functions.  The worst GD for MIES.ERT is approx. -100 for function 6 compared to the value of approx. -580 for function 14 in the case of (1+1)EA. This helps us make informed decisions for a given application.

 \begin{figure}[!t]
\centering
{{\includegraphics[width=0.8\textwidth]{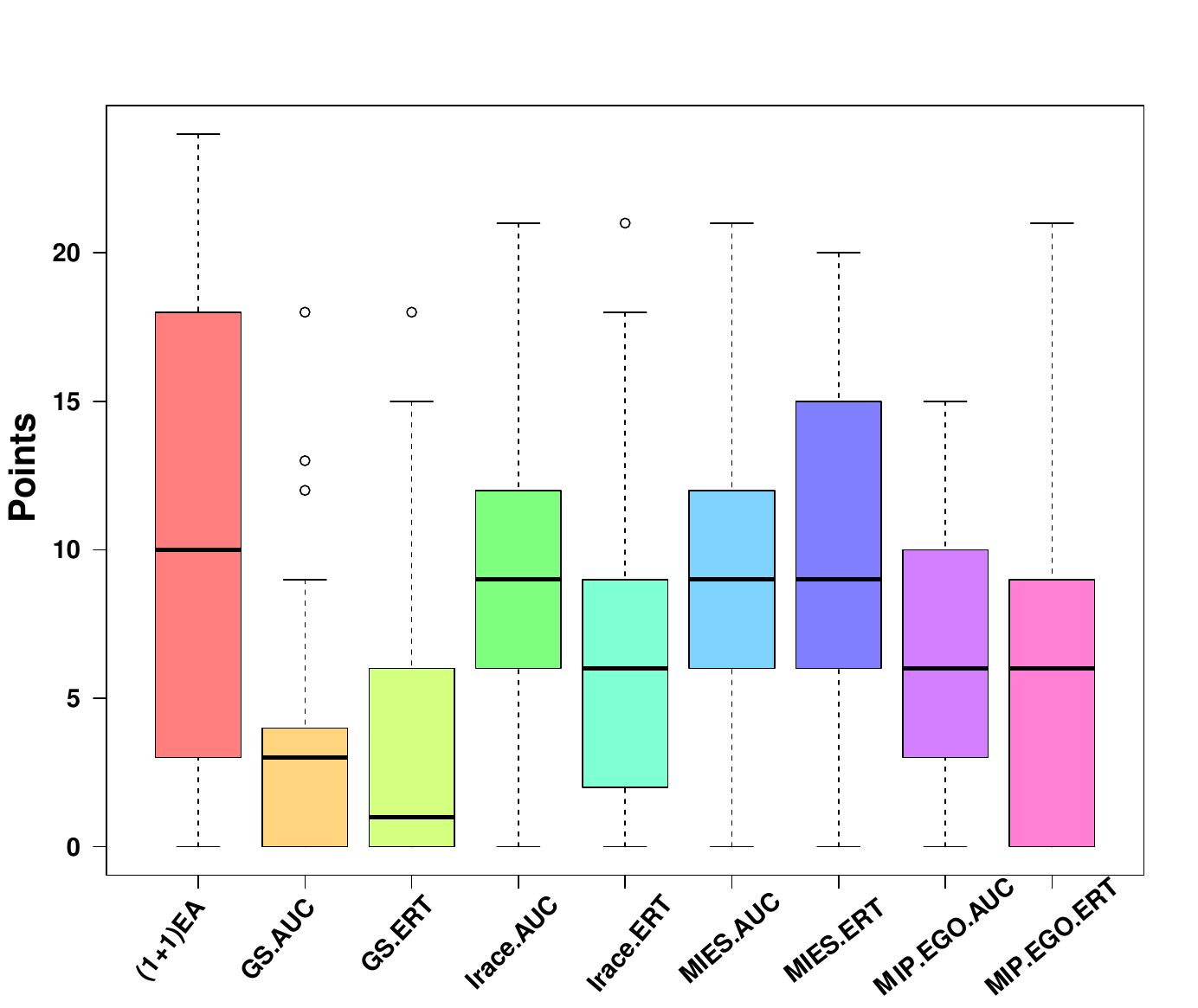} }}
    \caption{Proposed Ranking Scheme: Distribution of the \emph{points} attained by each algorithm for all 25 problems.}
\label{fig:bp}
  \end{figure} 
Considering page restrictions, function-wise ranking metrics is discussed only for the top 2 algorithms. The overall distribution of the \emph{points} for each algorithm is presented as a box plot in Figure \ref{fig:bp}. This also depicts the sequence of rankings as in Table \ref{tab:result_summary}. Considering the average mean, median and SD statistics of the \emph{points}, MIES.ERT, MIES.AUC and Irace.AUC exhibits similar performances. Also, the performances of Irace.ERT, MIP.EGO.ERT and MIP.EGO.AUC are comparable. 
The ranking produced by the classical boostrapped based HT is the same with the only exception of MIP.EGO.AUC outperforming MIP.EGO.ERT by one point. It is also evident that the performances of both algorithms in our proposed ranking scheme are comparable as in Figure \ref{fig:bp}. Again, this highlights the importance of $GD$ metric, which clearly showcases a very high positive goal difference for MIP.EGO.ERT and very large negative goal difference for MIP.EGO.AUC. This  explains the order that is provided by the proposed algorithm which gives a larger weight for practical significance (i.e., 3 \emph{points}). 
\begin{table}[!]
    \centering
\caption{\label{tab:result_summary} Proposed Ranking Scheme Results vs classical bootstrapped-based HT. The \emph{points} and GD obtained are the cumulative \emph{points} and GD obtained by each algorithm for all the 25 PBO problems.  The Change in ranking positions 
is indicated for classical HT with (↑, ↓).}
\begin{tabular}{|c|c|c|c|c|c|c|}
    \hline
    \multirow{2}{*}{Algorithm} & \multicolumn{2}{c|}{Proposed Ranking} & \multicolumn{2}{c|}{classical HT} 
    \\
    \cline{2-5}
& \emph{points} &  Goal Difference & \emph{points} & Change \\  
   \hline
1+1 EA & 259 & -755 & 107 & - \\ 
   \hline
MIES.ERT & 251 & 889 &103 & - \\
   \hline
MIES.AUC & 240 & 899 & 100 & -\\
   \hline
Irace.AUC & 227 &804 &89 & - \\
   \hline
Irace.ERT & 169 & -207 & 67 & -\\
   \hline
MIP.EGO.ERT & 161& 368 & 59 & $\downarrow$ 1\\
   \hline
MIP.EGO.AUC & 144 & -788 &60 & $\uparrow$ 1\\
   \hline
GS.ERT & 113& -544 & 41 & - \\
   \hline
GS.AUC & 91 & -820 & 35 & -\\
   \hline
\end{tabular}
\end{table}
\subsection{Sensitivity analysis of $\delta_p$ and $S$}
The sensitivity of input parameters on the ranking solutions are validated at desired severity levels of $S={0.5,0.65,0.8,0.95}$. Similarly, the practically relevant function evaluation is evaluated for a very wide range of $\delta_p={50,100,250,500}$, and the results of the ranking scheme for the resulting 16 experiments are compared. Table \ref{tab:result_sev} presents the influence of severity on the algorithm ranking for $\delta_p$ of 500. As expected, as the severity values increase from 50\% to 95\%, the test becomes more stringent and this is evident in the decreasing trend of \emph{points}. The order of the algorithm rankings is consistent even for various values of severity.  Note, however the magnitude of the change is a function of statistics of the solution of the algorithms and cannot be adjudged prior to the experiments.
Table \ref{tab:result_dels}, examines the influence of $\delta_p$ parameter on the outcome of the rankings. The order of the algorithm rankings remains consistent for $\delta_p$ ranging from 100 to 500. But when $\delta_p$ is 50, the MIP.EGO.AUC secures 178 \emph{points} with -7877 GD and is ahead of MIP.EGO.ERT which secured 177 \emph{points} with 3685 GD. This is indicated by change in positions ($\uparrow,\downarrow$). However, this is not a significant improvement for MIP.EGO.AUC, which is evident with the highly negative goal differences. It is to be noted that choosing extreme values for $\delta_p$  might influence the sequence of ranking and that is intentional. However, the comparison for all algorithms is performed using the same $\delta_{p}$ and the relative performance improvement is obtained.
\begin{table}[!]
    \centering
\caption{\label{tab:result_sev} Sensitivity analysis of parameter severity in Ranking Results ($\delta_p$=500). Given a $\delta_p$, the \emph{points} achieved by each algorithm decreases monotonically with  respect to the  increase in severity. As severity increases, the algorithms that pass a more stringent hypothesis test decreases and accordingly scores less number of \emph{points}. However, no prior statement can be given for the trend in GD for this case because it is a function of $\delta_p$. It is important to note that the sequence of the rankings of the algorithms remains unaltered with various severity levels.}
\begin{tabular}{|c|c|c|c|c|c|c|c|c|c|c}
    \hline
    \multirow{2}{*}{Algorithm} & \multicolumn{2}{c|}{$S$ 50\%} & \multicolumn{2}{c|}{$S$ 65\%} & \multicolumn{2}{c|}{$S$ 80\%} &\multicolumn{2}{c|}{$S$ 95\%} \\
    \cline{2-9}
    & \emph{points} & GD& \emph{points}& GD & \emph{points} & GD & \emph{points} & GD  \\
 1+1 EA & 271 & -855  & 263 & -806&259 & -755& 249 & -669\\ 
    \hline
 MIES.ERT & 257 & 952 &  251& 926& 251 & 889 &241 & 806\\
    \hline
MIES.AUC & 246 & 949 & 242& 930 &240 & 899 &232& 836 \\
    \hline
Irace.AUC & 229 & 864 & 227 & 834 & 227 &804 & 223 & 736\\
    \hline    
Irace.ERT & 171 & -205 & 171 & -205 &169 & -207 & 167 & -224\\
    \hline
 MIP.EGO.ERT & 165& 344 &161& 355&161& 368& 159 & 375 \\
    \hline
 MIP.EGO.AUC & 148 & -934 &146 & -865 & 144 & -788& 142 &-673 \\
    \hline
GS.ERT & 113& -608 & 113& -578 & 113& -544& 109 & -494 \\
    \hline
GS.AUC& 95 & -975 & 93 & -903 & 91 & -820& 91 & -693 \\
  \hline
\end{tabular}
\end{table}
\begin{table}[!]
    \centering
\caption{\label{tab:result_dels} Sensitivity analysis of parameter $\delta_p$ in Ranking Results ($S$=0.8). For a given value of severity, the GD decreases with the increase in the value of $\delta_p$ because of the inversely proportional relationship. This also holds true for the number of \emph{points} achieved.}
\begin{tabular}{|c|c|c|c|c|c|c|c|c|}
    \hline
    \multirow{2}{*}{Algorithm} & \multicolumn{2}{c|}{$\delta_p$=50} & \multicolumn{2}{c|}{$\delta_p$ = 100} & \multicolumn{2}{c|}{$\delta_p$=250} & \multicolumn{2}{c|}{$\delta_p$=500} \\
    \cline{2-9}
    & \emph{points} & GD& \emph{points}& GD & \emph{points}& GD & \emph{points}& GD \\
 1+1 EA & 319 & -7548 & 313& -3774 & 301 &-1510 & 259 & -755 \\ 
    \hline
MIES.ERT & 309 & 8890 &  307 & 4445& 285 & 1778 & 251 & 889\\
      \hline
MIES.AUC &  294 & 8995 & 290 & 4497  & 280 & 1799 & 240 & 899\\
    \hline
Irace.AUC & 265 & 8041 & 255 & 4020 &  245&1608& 227 & 804\\
    \hline    
Irace.ERT & 195& -2067 & 187 & -1034 & 181 &-414 & 169 & -207\\
    \hline
MIP.EGO.ERT & $\downarrow$ 177 & 3685 & 175 & 1842 & 169 & 737 & 161 & 368 \\
    \hline
MIP.EGO.AUC& $\uparrow$ 178 & -7877 & 172 & -3939 & 156 & -1576 & 144 & -788 \\
    \hline
GS.ERT & 123&-5434& 121&-2717 & 115& -1087 & 113& -544\\
    \hline
GS.AUC& 105& -8194 & 105& -4097& 99& -1639 & 91&-820\\
  \hline
\end{tabular}
\end{table}
 Even in the existing ranking schemes, the practically relevant improvement parameter is defined as a user-defined variable \cite{efti17a,efti19b,efti19a,Niki14a}, and extreme choice of this parameter will alter the results. Considering the page restrictions, only certain experiments are presented in Table \ref{tab:result_sev} and Table \ref{tab:result_dels}. In addition, repeating the same experiment for several times always produced the same results, convincing of the robust and stable outcome of the proposed ranking scheme.
The CRS4EAs ranking scheme as proposed in \cite{Niki14a}, is used to compare the results obtained by our proposed ranking scheme. The results of the same 9 algorithms for 25 PBO problems are provided as input and in total 5000 games were played. The implementation available in \cite{ioh22a} is used for the experiments with the default suggested input parameter settings. The results are shown in Table \ref{tab:result_chess}. It is evident that for the same input, the resulting ranking and the ratings keep varying when the same experiment is repeated. This is indicated by the change in positions ($\uparrow and \downarrow$). Hence, consistent ranking is not observed even with the same inputs. 
\begin{table}[!]
    \centering
\caption{\label{tab:result_chess} Results of CRS4EAs ranking scheme for 3 random runs but for the same input settings. The algorithms are ranked as per the proposed ranking scheme and the change in rankings based on CRS4EAs is indicated.}
\begin{tabular}{|c|c|c|c|c|c|c|c|c|c|c|c|c|}
    \hline
    \multirow{2}{*}{Algorithm} & \multicolumn{4}{c|}{Trial 1} & \multicolumn{4}{c|}{Trial 2} & \multicolumn{4}{c|}{Trial 3}  \\
    \cline{2-13}
    & R & RD & $\sigma$ &Change&  R & RD & $\sigma$ &Change& R & RD & $\sigma$ &Change \\
 1+1 EA & 1586&12.7&0.04& -& 1578& 12.7& 0.04&$\downarrow$1& 1566& 13.1&0.04 &$\downarrow$2\\ 
    \hline
MIES.ERT & 1569&13.5&0.04 &$\downarrow$1&1574&13.8&0.04&$\downarrow$1&1575&13.3&0.04& -\\
    \hline
MIES.AUC & 1586& 14.9& 0.06 &$\uparrow$1&1604&14&0.05&$\uparrow$2& 1578&13.8&0.05&$\uparrow$2\\
    \hline
Irace.AUC & 1563& 13.1&0.04&-&1559&14.7&0.05&-&1559&12.9&0.04&-\\
    \hline  
Irace.ERT & 1457&13.5&0.04 & $\downarrow$2& 1451&12.7&0.04&$\downarrow$2& 1436&13.6&0.04&$\downarrow$2\\
    \hline
 MIP.EGO.ERT & 1479&12.8&0.04 &$\uparrow$1& 1495&13.5&0.04&$\uparrow$1&1483&12.2&0.03&- \\
    \hline
 MIP.EGO.AUC & 1473& 13.4& 0.04 & $\uparrow$ 1&1480&13.1&0.04& $\uparrow$ 1& 1489&13.2&0.04&$\uparrow$ 2\\
    \hline
  GS.ERT & 1377&14.2&0.05 &$\downarrow$ 1& 1371&14.2&0.05&$\downarrow$ 1 &1398&13.8&0.04&$\downarrow$ 1 \\
    \hline
GS.AUC & 1410& 13.1& 0.04& $\uparrow$ 1 & 1400 & 13.9& 0.04& $\uparrow$ 1 & 1415& 14.1& 0.05& $\uparrow$ 1 \\
  \hline
\end{tabular}
\end{table}
\section{Summary and Outlook}\label{sec:summary}
A user-friendly novel ranking scheme based on football league system is proposed that takes into account the statistical significance, the practical significance, and the magnitude of the win or loss of the compared algorithms to determine the final ranking of the algorithms. The proposed scheme has the advantage that the order of comparison has no impact on the results along with there is no necessity of knowing the prior statistics of the compared algorithms. Since the practical significance is used separately from hypothesis testing, the resulting scheme does not make the test more conservative. The proposed scheme shows potential also for comparing machine learning, artificial intelligent(AI), and explainable AI algorithms. For didactic purposes, the HT is formulated considering the mean performance differences among algorithms. Nevertheless, the scheme can be used to compare the median and related performance measures as well.
%
%
\newpage
 \bibliographystyle{splncs04}
 \bibliography{chandrasekaran}
\end{document}